\documentclass[journal]{IEEEtran}
\usepackage{geometry}
\usepackage[utf8]{inputenc}
\usepackage{amsmath}
\usepackage{tabulary}
\usepackage{tabularx}
\usepackage{adjustbox}
\usepackage{url}
\usepackage{hyperref}
\usepackage{amsmath,amssymb,amsfonts}
\usepackage{algorithm}
\usepackage{algpseudocode}
\usepackage{graphicx}
\usepackage{textcomp}
\usepackage{xcolor}
\usepackage{multirow}
\usepackage{booktabs}
\begin{document}

\title{An Investigation into Seasonal Variations in Energy Forecasting for Student Residences}

\author{
    \IEEEauthorblockN{Muhammad Umair Danish\IEEEauthorrefmark{1},
    Mathumitha Sureshkumar\IEEEauthorrefmark{2},
    Tehara Fonseka\IEEEauthorrefmark{3},
    Umeshika Uthayakumar\IEEEauthorrefmark{4},
    and Vinura Galwaduge\IEEEauthorrefmark{5}}
    
    \IEEEauthorblockA{Department of Electrical and Computer Engineering\\
    Western University, London, Ontario, Canada\\
    \IEEEauthorrefmark{1}mdanish3@uwo.ca,
    \IEEEauthorrefmark{2}msureshk@uwo.ca,
    \IEEEauthorrefmark{3}tfonsek@uwo.ca,
    \IEEEauthorrefmark{4}uuthayak@uwo.ca,
    \IEEEauthorrefmark{5}vgalwadu@uwo.ca\\
    \textit{All authors contributed equally to this work.}}
}
\maketitle

\begin{abstract}
This research provides an in-depth evaluation of various machine learning models for energy forecasting, focusing on the unique challenges of seasonal variations in student residential settings. The study assesses the performance of baseline models, such as LSTM and GRU, alongside state-of-the-art forecasting methods, including Autoregressive Feedforward Neural Networks, Transformers, and hybrid approaches. Special attention is given to predicting energy consumption amidst challenges like seasonal patterns, vacations, meteorological changes, and irregular human activities that cause sudden fluctuations in usage. The findings reveal that no single model consistently outperforms others across all seasons, emphasizing the need for season-specific model selection or tailored designs. Notably, the proposed Hyper Network based LSTM and MiniAutoEncXGBoost models exhibit strong adaptability to seasonal variations, effectively capturing abrupt changes in energy consumption during summer months. This study advances the energy forecasting field by emphasizing the critical role of seasonal dynamics and model-specific behavior in achieving accurate predictions.
\end{abstract}
\begin{IEEEkeywords}
MiniAutoEncXGBoost, sequential, long short-term memory, attention, transformer, hypernetwork, neural basis expansion, auto-regressive, MiniRocket, XGBoost, autoencoder  
\end{IEEEkeywords}

\section{Introduction}
The Independent Electricity System Operator (IESO) projects that Ontario's energy demand will grow exponentially over the next decade \cite{ieso2021}. Furthermore, the World Energy Council (WEC) emphasizes that energy efficiency will be critical in preventing scenarios where demand surpasses supply \cite{worldenergyreport}. Accurate energy forecasting is a pivotal component of energy planning and procurement processes; however, it has historically been a weakness in Ontario \cite{on360policy}. The primary objective of forecasting in the energy sector is to predict electricity demand and consumption to ensure an optimal supply plan. Inaccurate forecasts can lead to an over-supply or a shortage, resulting in financial and energy-related consequences.

Improving energy forecasting aligns closely with sustainability goals and yields significant financial savings. It also facilitates efficient energy management, optimized deployment and operational strategies, informed infrastructure planning, reduced carbon footprints, and minimized costs. The challenges of accurate energy forecasting arise from the inability of some models to capture energy patterns or achieve convergence effectively. Furthermore, energy consumption predictions are profoundly influenced by the unpredictable nature of human behavior and natural events. A notable example is the COVID-19 pandemic, which significantly disrupted global energy consumption patterns due to its unprecedented impact on human behavior \cite{abulibdeh2022impact}.

This paper investigates 13 Machine Learning (ML) models, including introducing two novel models to predict day-ahead energy consumption using two real-world datasets: Residence 1 and Residence 2, located at Western University. The analysis aims to uncover specific energy usage patterns, anomalies, and limitations of predictive models influenced by unpredictable human behavior and external factors. The primary objective is to develop a robust short-term or day-ahead load forecasting model capable of significantly reducing forecasting errors in energy applications.

Initially, five baseline models were evaluated: Multi-Layer Perceptron (MLP), Temporal Convolution Neural Network (TCN), and three sequential neural networks—Recurrent Neural Network (RNN), Gated Recurrent Unit (GRU), and Long Short-Term Memory (LSTM). Subsequently, the study explored state-of-the-art and ensemble models to assess their potential for improved accuracy and adaptability across seasons. These models included Neural Basis Expansion Analysis for Interpretable Time Series Forecasting (N-BEATS) \cite{oreshkin2019n}, Auto-Regressive Feedforward Neural Network (AR-Net) \cite{triebe2019ar}, and a hybrid of MiniRocket, AutoEncoder, and XGBoost. Additionally, the performance of Transformer models and an Integration of Attention Mechanisms with LSTM was examined. A Hypernetwork model was designed and integrated with an LSTM network to enhance forecasting robustness further.

The remainder of this report is organized as follows. Section II provides the background on the employed algorithms and accuracy measures. Section III reviews related works in this domain. Section IV details the methodology, including data preprocessing, feature engineering, and the validation process. Section V presents the evaluation metrics, results, and analysis. Finally, Section VI concludes the report and discusses potential future work.

\section{BACKGROUND}
This section briefly describes the background information of the project including the model architectures used for training, and the metrics used to evaluate the performance.
\subsection{Model Architectures}
We began the project by exploring five baseline model architectures: Multi-Layer Perceptron (MLP), Temporal Convolution, Recurrent Neural Network (RNN), Long Short-Term Memory (LSTM), and Gated Recurrent Unit (GRU). These models served as foundational benchmarks for our energy forecasting task. MLP, as one of the most straightforward deep learning architectures, is often employed for time series prediction tasks. Temporal Convolutions, on the other hand, leverage 1-D convolutional layers to efficiently capture temporal dependencies in sequential data \cite{lara2020temporal}. 

The remaining three baseline models—RNN, LSTM, and GRU—were chosen for their specialization in sequence modeling. Additional feature extraction techniques were integrated additional feature extraction techniques were integrated to enhance the performance of these baseline models. These included temporal convolution-based methods such as MiniRocket \cite{dempster2021minirocket}, a highly efficient approach for feature extraction, and a temporal convolution-based autoencoder architecture. Furthermore, an LSTM-based architecture incorporating a self-attention layer was implemented to better capture long-range dependencies in time series data.

In addition to these baseline models, we experimented with two advanced architectures: Neural Basis Expansion Analysis for Interpretable Time Series Forecasting (N-BEATS) and Auto-Regressive Feedforward Neural Network (AR-Net), both introduced in recent literature. The N-BEATS model utilizes fully connected layers organized into multiple building blocks with forward and backward residual links, enabling interpretable and robust forecasting \cite{oreshkin2019n}. AR-Net combines autoregressive statistical modeling with feedforward neural networks to bridge traditional statistical methods and modern deep learning approaches \cite{triebe2019ar}.

\subsection{Evaluation Metrics}
To assess the performance of the models on the energy forecasting task, we employed three metrics: Symmetric Mean Absolute Percentage Error (SMAPE), Mean Absolute Error (MAE), and Root Mean Square Error (RMSE). While Mean Absolute Percentage Error (MAPE) is commonly used due to its interpretability, we opted for SMAPE as it is scale-independent and less sensitive to the magnitude of forecasted values. This characteristic is particularly advantageous for time series data with varying scales. The formula for SMAPE is:
\begin{equation}
    \text{SMAPE} = 100\% \times \frac{1}{n} \sum_{i=1}^n \frac{2 |y_i - \hat{y}_i|}{|y_i| + |\hat{y}_i|},
\end{equation}

MAE, which measures the average magnitude of errors without considering their direction, is calculated as follows:
\begin{equation}
    \text{MAE} = \frac{1}{n} \sum_{i=1}^{n} |y_i - \hat{y}_i|
\end{equation}

RMSE, which penalizes larger errors more heavily, is computed as:
\begin{equation}
    \text{RMSE} = \sqrt{\frac{1}{n} \sum_{i=1}^{n} (y_i - \hat{y}_i)^2}
\end{equation}

In these equations, \(y_i\) represents the actual energy value, \(\hat{y}_i\) denotes the predicted energy value, and \(n\) is the total number of observations. Together, these metrics provide a comprehensive evaluation of the forecasting performance.

\section{Related Work}

The field of energy forecasting encompasses a variety of approaches, each with its challenges. Traditional statistical models, such as ARIMA and State Space Models, often struggle to capture energy data's complex and nonlinear characteristics \cite{debnath2018forecasting}. Recurrent Neural Networks (RNNs) offer an improvement by capturing temporal sequences in energy consumption patterns. However, they are hindered by vanishing and exploding gradient problems, limiting their ability to model long-term dependencies crucial for accurate energy predictions. Extended Short-Term Memory Networks (LSTMs) address these issues through a sophisticated gating mechanism; however, this comes at the cost of increased computational complexity \cite{natarajan2019survey}.

Recently, Transformer models, initially proposed for natural language processing tasks, have gained popularity in energy forecasting. Introduced by Vaswani et al. \cite{vaswani2017attention}, Transformers revolutionized sequential data processing by eliminating the reliance on sequential input order. Instead, they leverage self-attention mechanisms to weigh the significance of different parts of the input data, making them particularly effective in handling complex energy consumption patterns. A study by L'Heureux et al. \cite{l2022transformer} highlights the effectiveness of Transformer-based models in electrical load forecasting, demonstrating their potential for more accurate and computationally efficient predictions than traditional models.

Tree-based models, such as Gradient Boosted Decision Trees (GBDTs) and XGBoost, have also been widely used in load forecasting tasks. While these models excel in general predictive tasks, they face challenges in capturing complex transient patterns, such as sudden spikes in energy consumption \cite{zhu2021hybrid}.

Feature extraction techniques are critical in improving model performance by creating robust representations of time-series data. Tools like Time Series Feature Extraction Library (TSFEL) \cite{barandas2020tsfel} and methods such as Discrete Wavelet Transformation (DWT) \cite{morchen2003time} have been applied to extract meaningful features from energy data. Additionally, convolutional autoencoders combined with LSTMs have been explored for tasks like unsupervised clustering in electricity consumption, further enhancing the representation of temporal data \cite{ryu2018residential}.

Recent advancements such as MiniRocket, a deterministic and efficient feature extraction method, have demonstrated competitive performance with fast feature computation, making them promising for energy forecasting applications \cite{bondugula2023novel}. However, computational costs remain a significant challenge in scenarios requiring rapid processing of large datasets.

Despite these challenges, advancements in energy forecasting models continue to provide valuable insights and methodologies. Integrating novel architectures and feature extraction techniques holds promise for addressing the complexities of energy forecasting and improving its accuracy and efficiency.

\section{Methodology}
This section provides an overview of the datasets, preprocessing techniques, and implemented models. It includes descriptions of the baseline models, state-of-the-art approaches, hybrid methodologies, and the proposed models.

\subsection{Dataset}
\begin{figure}[H]
    \centering
    \includegraphics[width=0.99\linewidth]{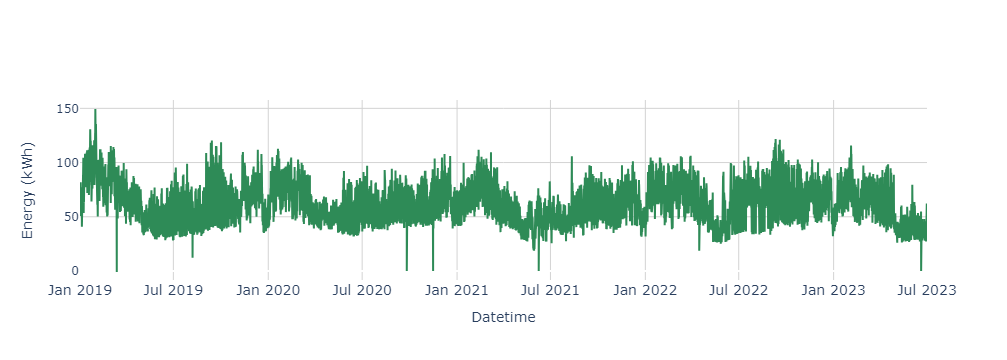}
    \caption{Energy consumption over three years for Student Residence 1, illustrating the variability and usage patterns within this specific student residential setting.}
    \label{fig:essex_energy}
\end{figure}

\begin{figure}[H]
    \centering
    \includegraphics[width=0.99\linewidth]{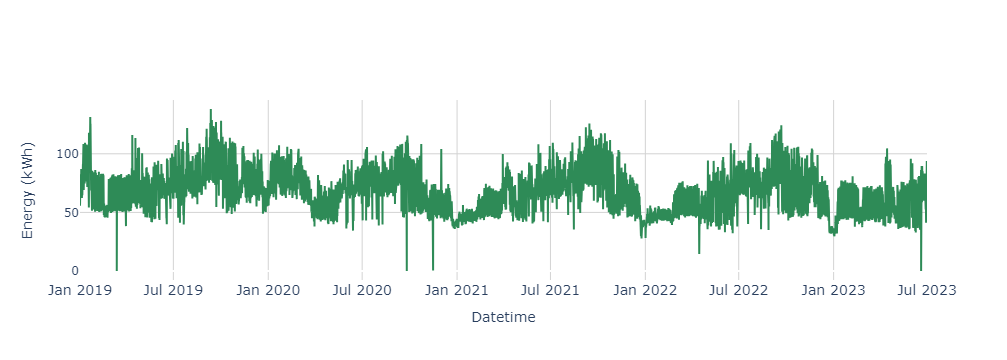}
    \caption{Energy consumption over three years for Student Residence 2, illustrating the variability and usage patterns within this specific student residential setting.}
    \label{fig:perth_energy}
\end{figure}

We utilized two real-world datasets capturing electricity consumption (in kWh) of two residence buildings (Residence-1 and Residence-2) at Western University in London, Ontario, spanning January 2019 to June 2023. Figures \ref{fig:essex_energy} and \ref{fig:perth_energy} illustrate these datasets' energy consumption trends and seasonality. Residence 1 exhibited more erratic consumption patterns than Residence 2, but neither dataset displayed distinct cyclic irregularities; instead, they contained a mix of seasonal and trend components.

\subsubsection{Feature Enhancement}
The original dataset consisted of three features: date-time, temperature, and energy (in kilowatt-hours). Since neural networks cannot interpret date-time information in text format, additional features were derived from the date-time attribute. These included the day of the year, the day of the month, the day of the week, and the hour.

\subsubsection{Selected Features}
Six features were used for all models: day of the year, day of the month, day of the week, hour, temperature, and energy.

\subsubsection{Data Preprocessing Techniques}
\paragraph{MinMax Scaling}
MinMax scaling was applied to normalize the dataset. This technique is widely used in energy forecasting due to its efficiency and minimal sensitivity to outliers. The formula for MinMax scaling is:
\begin{equation}
X_{\text{scaled}} = \frac{X - X_{\text{min}}}{X_{\text{max}} - X_{\text{min}}}
\end{equation}
where \( X \) is the original data, \( X_{\text{min}} \) and \( X_{\text{max}} \) are the minimum and maximum values in the dataset, and \( X_{\text{scaled}} \) is the normalized data.

\paragraph{Sliding Window Technique}
A sliding window technique was used to create temporal data windows. Each input window, \( X_{24} \), represents the preceding 24 hours of data, including energy, temperature, day of the year, day of the month, day of the week, and hour. The target window, \( \hat{y}_{24} \), corresponds to the next 24 hours of energy consumption. A stride of one was employed to capture patterns effectively.

The sliding window operation can be expressed as:
\begin{equation}
X_{24}, \hat{y}_{24} = \mathcal{SW}(data, 24, 1)
\end{equation}
where \( data \) represents the complete dataset, `24` is the window size, and `1` is the stride. \(\mathcal{SW}\) denotes the sliding window operation.

\subsubsection{Data Splitting Strategy}
The dataset was partitioned into training, validation, and test sets, with 70\% allocated for training, 10\% for validation, and the remaining 20\% for testing.

\subsection{Energy Forecasting with Basic Models}
The initial phase of this study involved training baseline models to forecast energy consumption and using their results as a benchmark for comparison with advanced approaches. To optimize the performance of these models, hyperparameters such as the number of neurons in hidden layers and dropout rates were tuned. Additional strategies like dropout layers and early stopping were employed during training to mitigate overfitting.

The MLP model consisted of two fully connected layers, with a dropout layer preceding the output layer. The Temporal Convolution model included three 1D-convolution layers interspersed with two dropout layers. The sequential models (RNN, LSTM, and GRU) shared a similar architecture: a single sequential module followed by a dropout layer and a fully connected output layer.

\begin{figure}[H]
    \centering
    \includegraphics[width=0.7\linewidth]{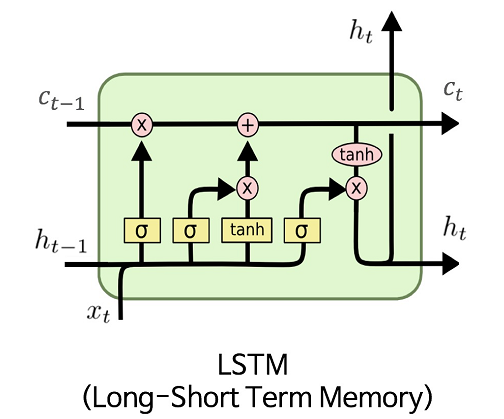}
    \caption{Architecture of one of the baseline models: LSTM, a standard model used in energy forecasting.}
    \label{fig:lstm}
\end{figure}

The process can be mathematically described as:
\begin{equation}
\hat{y_{24}} = \mathcal{FC}(\mathcal{M}(X_{24}))
\end{equation}
Where \( \hat{y_{24}} \) represents predictions for the next 24 hours, \( \mathcal{M}(X_{24}) \) denotes the processing of input data by the baseline model \( \mathcal{M} \), and \( \mathcal{FC} \) is a fully connected layer that generates the final prediction.

\subsection{Energy Forecasting with State-of-the-Art Models}
We employed two state-of-the-art models: N-BEATS and AR-Net.

\begin{figure}[H]
    \centering
    \includegraphics[width=0.7\linewidth]{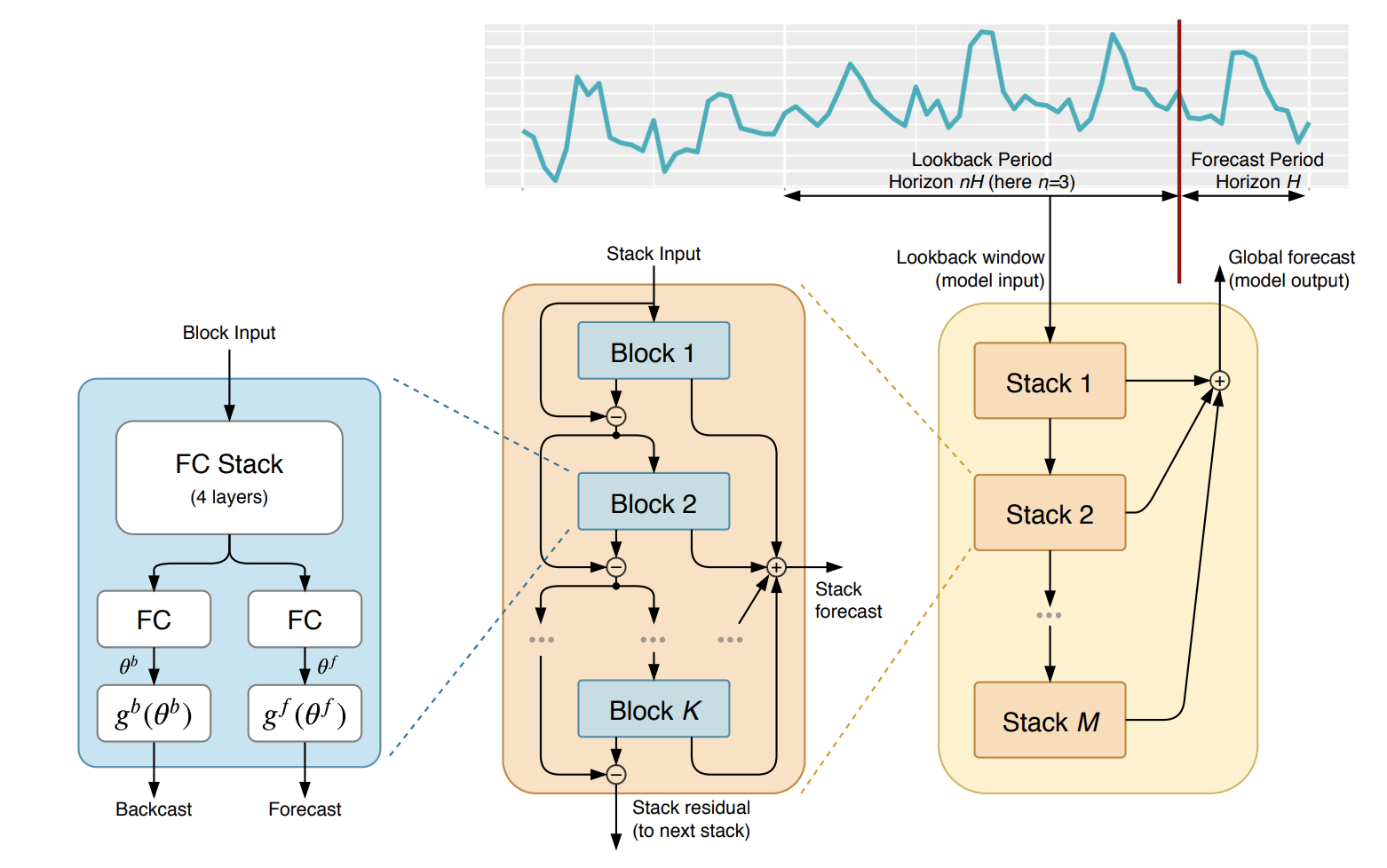}
    \caption{The Neural Basis Expansion Analysis model is an MLP-based architecture consisting of layers, stacks, and blocks.}
    \label{fig:nbeats}
\end{figure}

The N-BEATS model is based on an ensemble of stacked blocks, where each block comprises feed-forward neural networks with ReLU activation functions. These blocks predict basis expansion coefficients aggregated to form the final forecast. This architecture effectively captures time dependencies and patterns in time-series data \cite{oreshkin2019n}. In our implementation, six N-BEATS blocks with four hidden layers were utilized. Figure \ref{fig:nbeats} illustrates the N-BEATS architecture from the original study.

\begin{figure}[H]
    \centering
    \includegraphics[width=0.7\linewidth]{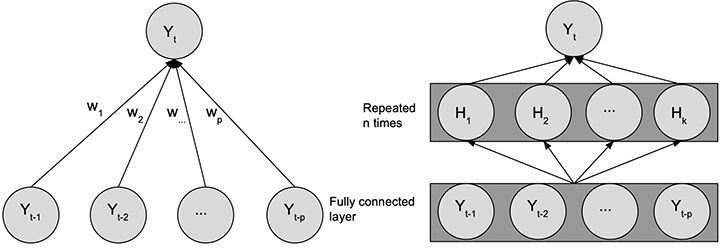}
    \caption{Left: Basic AR-Net without hidden layers. Right: Advanced AR-Net with multiple hidden layers.}
    \label{fig:arnet}
\end{figure}

The AR-Net model integrates auto-regressive coefficients as parameters of its first layer, alongside linear layers and ReLU activation functions \cite{triebe2019ar}. This model was designed with two hidden layers to evaluate the precision of fitted auto-regressive coefficients. Figure \ref{fig:arnet} illustrates its architecture.

\subsection{Transformer}
Transformers, introduced by Vaswani et al. \cite{vaswani2017attention}, have revolutionized sequential data modeling. Their architecture incorporates positional encoding, multi-head self-attention, and feed-forward layers. For this study, only the encoder part of the Transformer was utilized due to data constraints. 

\begin{figure}[H]
    \centering
    \includegraphics[width=0.7\linewidth]{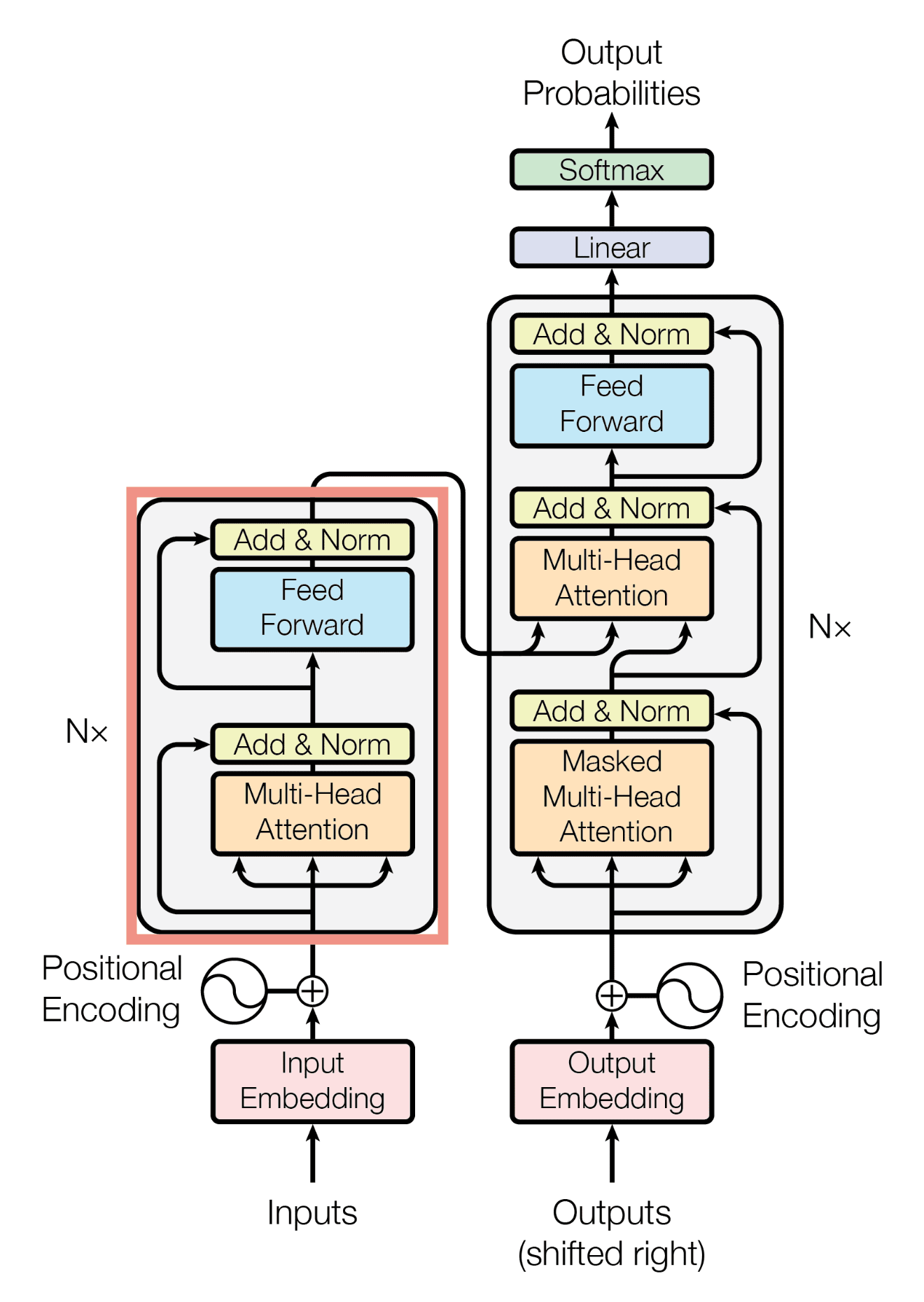}
    \caption{The encoder-based Transformer architecture with fully connected layers before and after the encoder.}
    \label{fig:transfromer}
\end{figure}

The process is mathematically expressed as:
\begin{equation}
\hat{y_{24}} = \mathcal{FC}(\mathcal{TE}(X_{24}))
\end{equation}
Where \( \hat{y_{24}} \) denotes predictions for the next 24 hours, \( \mathcal{TE}(X_{24}) \) represents the processing of input data by the Transformer encoder \( \mathcal{TE} \), followed by a fully connected layer \( \mathcal{FC} \) for the final prediction.

\subsection{Integration of Attention and LSTM}
A self-attention layer was integrated with an LSTM layer to enhance long-range dependency modeling. The architecture consists of an LSTM layer, a self-attention layer, and a fully connected layer for predictions. An optional dropout layer was included for regularization.

\begin{figure}[H]
    \centering
    \includegraphics[width=0.9\linewidth]{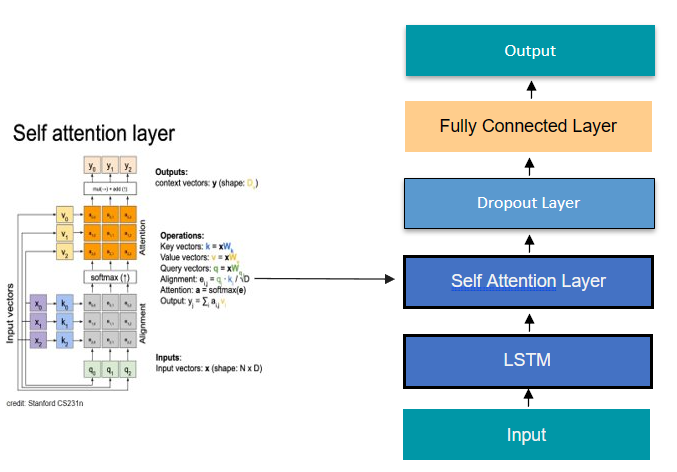}
    \caption{The architecture combines a self-attention layer and LSTM for energy forecasting.}
    \label{fig:attention}
\end{figure}

The process is mathematically represented as follows:
\begin{equation}
\hat{y_{24}} = \mathcal{FC}(\mathcal{SA}(\mathcal{LSTM}(X_{24})))
\end{equation}
where \( \mathcal{SA} \) represents the self-attention layer applied after the LSTM.

\subsection{Ensemble Models for Energy Forecasting}
Ensemble models were developed by combining feature extraction techniques. Two baseline ensemble models were built:
\begin{enumerate}
    \item MiniRocket and Stochastic Gradient Descent (SGD) Regressor: Features extracted by MiniRocket were fed into an SGD regressor with a multi-output wrapper \cite{dempster2021minirocket}.
    \item MiniRocket and XGBoost: Features from MiniRocket were used as input to an XGBoost model.
\end{enumerate}

Further, a novel ensemble was designed, combining MiniRocket and a convolutional autoencoder for feature extraction, with XGBoost as the final regression layer. This approach leverages the strengths of both feature extractors and tree-based prediction techniques.

\subsection{Proposed Models}

\paragraph{\textbf{Hypernetwork and LSTM}} 
We propose an energy forecasting architecture combining a primary LSTM network with a Hypernetwork. The Hypernetwork dynamically generates weights for the LSTM based on input features, enabling adaptive behavior. The Hypernetwork consists of three fully connected layers with ReLU activation functions for non-linearity and Xavier for weight initialization. The primary network comprises a standard LSTM layer and a fully connected layer for final predictions.

\begin{figure}[H]
    \centering
    \includegraphics[width=0.9\linewidth]{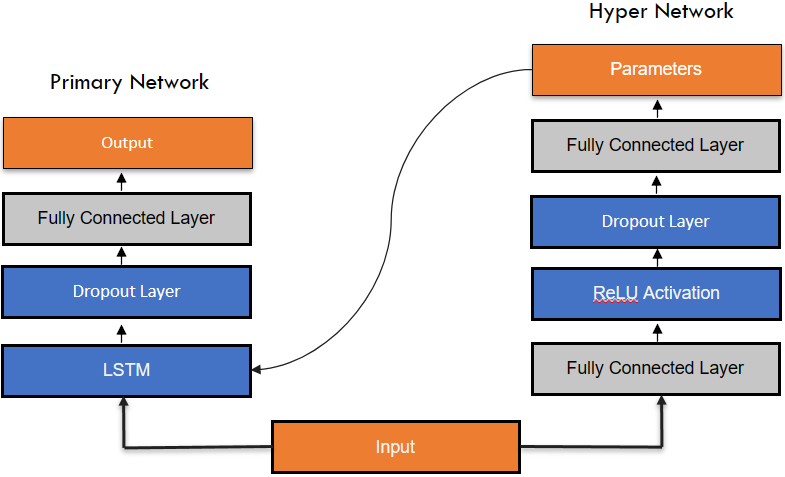}
    \caption{The architecture of the Hypernetwork-based MLP and primary network-based LSTM.}
    \label{fig:attention1}
\end{figure}

The Hypernetwork process is mathematically represented as:
\begin{equation}
\theta_{\mathcal{HN}} = \mathcal{FC}_{\mathcal{HN}}(X_{24})
\end{equation}
where \( \theta_{\mathcal{HN}} \) are the dynamically generated weights for the LSTM, and \( \mathcal{FC}_{\mathcal{HN}} \) represents the fully connected layers of the Hypernetwork.

The primary LSTM network's process is given by:
\begin{equation}
\hat{y_{24}} = \mathcal{FC}(\mathcal{LSTM}(X_{24}; \theta_{\mathcal{HN}}))
\end{equation}
where \( \mathcal{LSTM}(X_{24}; \theta_{\mathcal{HN}}) \) processes the input data \( X_{24} \) with the weights \( \theta_{\mathcal{HN}} \) generated by the Hypernetwork, and \( \mathcal{FC} \) is a fully connected layer producing the final output \( \hat{y_{24}} \).

\paragraph{\textbf{Ensemble of MiniRocket, Convolutional Autoencoder, and XGBoost}} 
This ensemble model leverages two feature extractors, MiniRocket, and a Convolutional Autoencoder, combined with XGBoost for energy forecasting.

\begin{figure}[H]
    \centering
    \includegraphics[width=0.9\linewidth]{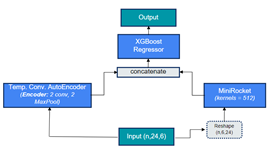}
    \caption{The ensemble model with a convolutional autoencoder and MiniRocket as feature extractors.}
    \label{fig:ensemble}
\end{figure}

The convolutional autoencoder uses a simple encoder architecture with two 1D convolutional layers interspersed with max-pooling layers. The decoder mirrors the encoder with deconvolution and up-sampling layers. The input data \( X_{24} \) is reshaped (from \( (n,24,6) \) to \( (n,6,24) \), where \( n \) is the batch size) before being fed into the MiniRocket model to perform convolution operations along a different dimension, extracting diverse features. The autoencoder encoder architecture is mathematically described as follows:

\begin{equation}
y = ReLU(Conv1D(X_{24}))
\end{equation}
where \( \text{num\_kernels} = 16, \text{kernel\_size} = 2 \).

\begin{equation}
y = MaxPool1D(y)
\end{equation}

\begin{equation}
y = ReLU(Conv1D(y))
\end{equation}
where \( \text{num\_kernels} = 8, \text{kernel\_size} = 3 \).

\begin{equation}
y_1 = Flatten(MaxPool1D(y))
\end{equation}

The final ensemble architecture combines the outputs of the autoencoder \( \mathcal{AE} \), MiniRocket \( \mathcal{MR} \), and XGBoost \( \mathcal{XGB} \) as follows:
\begin{equation}
y_1 = \mathcal{AE}(X_{24})
\end{equation}
\begin{equation}
y_2 = \mathcal{MR}(\text{reshape}(X_{24}))
\end{equation}
\begin{equation}
\hat{y_{24}} = \mathcal{XGB}(\text{concat}(y_1, y_2))
\end{equation}

This ensemble model effectively balances computational efficiency with predictive accuracy. MiniRocket, for instance, can extract features from the training dataset (\( (55077, 24, 6) \)) in approximately 47 seconds using a single Intel® Xeon® CPU, making it a practical solution for large-scale energy forecasting tasks.

\section{Results and Evaluation}
This section outlines the hyperparameter optimization methodology, defines the objective functions, discusses regularization techniques, and presents analyses of overall results and seasonal performance evaluations.

\subsection{Hyperparameter Optimization}
All models were trained on the same dataset, maintaining input and output dimensions consistency. A Grid Search approach was employed to identify the optimal configurations of units, layers, optimizers, and learning rates for each model. Table \ref{tab:search_space_transformer} details the search space for optimizing the Transformer model hyperparameters.

\begin{table}[H]
\caption{Search Space for Transformer Model Hyperparameters}
\centering
\begin{tabular}{|l|l|}
\hline
\textbf{Parameter}                     & \textbf{Possible Values}         \\ \hline
Number of Encoder Layers               & 1, 2, 4, 6                       \\ \hline
Dimension of Feedforward Network       & 128, 256, 512                    \\ \hline
Number of Attention Heads              & 2, 4, 6                          \\ \hline
Learning Rate                          & 0.1, 0.001, 0.004                \\ \hline
Optimizer                              & Adam, SGD, AdamW                 \\ \hline
Loss Function                          & MSE, MAE                         \\ \hline
\end{tabular}
\label{tab:search_space_transformer}
\end{table}

\subsection{Objective Function}
The primary objective for all models was to minimize the difference between actual and predicted values. Two loss functions were employed:

\paragraph{Mean Squared Error (MSE):}
\begin{equation}
\text{MSE} = \frac{1}{24} \sum_{i=1}^{24} (y_i - \hat{y}_i)^2
\end{equation}

\paragraph{Mean Absolute Error (MAE):}
\begin{equation}
\text{MAE} = \frac{1}{24} \sum_{i=1}^{24} |y_i - \hat{y}_i|
\end{equation}
where \(y_i\) is the actual energy value, \(\hat{y}_i\) is the predicted energy value, and \(24\) represents the number of observations. MSE was employed for most models, while MAE was specifically used for the Temporal Convolutional Network due to its superior performance with this metric.

\subsection{Regularization}
Weight decay was implemented as a regularization technique across all models to prevent overfitting. This penalizes large weight values, effectively limiting model complexity.

\subsection{Optimized Hyperparameters for Base Models}
The optimized hyperparameters for the baseline models were obtained via Grid Search:
\begin{itemize}
    \item \textbf{GRU:} 64 units, the dropout rate of 0.1, the learning rate of 0.004.
    \item \textbf{LSTM:} 64 units, the dropout rate of 0.5, the learning rate of 0.001.
    \item \textbf{RNN:} 64 units, the dropout rate of 0.5, the learning rate of 0.01.
    \item \textbf{MLP:} 32 units, the learning rate of 0.004.
    \item \textbf{Temporal Convolutional Model:} 128 units, the dropout rate of 0.5, the learning rate of 0.01.
\end{itemize}
The Stochastic Gradient Descent (SGD) optimizer was employed for all baseline models.

\subsection{Optimized Hyperparameters for SOTA Models}
\begin{itemize}
    \item \textbf{N-BEATS:} 6 blocks, each with four hidden layers and 512 units.
    \item \textbf{ARFNN:} 64 units, two layers, learning rate of 0.001.
\end{itemize}
For both models, the SGD optimizer was the most effective.

\subsection{Optimized Hyperparameters for the Transformer Model}
The Transformer model configuration included the following:
\begin{itemize}
    \item 2 attention heads.
    \item Fully connected layers with 64 units before and 24 units after the encoder.
    \item Dropout rate of 0.1, learning rate of 0.001, and SGD optimizer.
\end{itemize}

\subsection{Optimized Hyperparameters for the Integration of Attention and LSTM Model}
The model was configured with:
\begin{itemize}
    \item LSTM layer: 64 units, dropout rate of 0.5.
    \item Learning rate: 0.01.
    \item Optimizer: Adam.
\end{itemize}

\subsection{Optimized Hyperparameters for the Hypernetwork and LSTM Model}
The optimal configuration was:
\begin{itemize}
    \item LSTM layer: 64 units, dropout rate of 0.1, learning rate of 0.01.
    \item Hypernetwork: 128 units in the first layer and 64 in the second layer.
    \item Optimizer: SGD.
\end{itemize}

\subsection{Optimized Hyperparameters for the Ensemble Model}
The ensemble model was optimized as follows:
\begin{itemize}
    \item \textbf{MiniRocket:} 512 convolution kernels, resulting in feature vectors of length 512.
    \item \textbf{Autoencoder:} AdamW optimizer, learning rate 0.0001, Mean Squared Error Loss.
    \item \textbf{XGBoost:} Learning rate of 0.05, colsample\_bytree of 0.5, reg\_lambda of 1.2, a subsample of 0.8, booster set to 'gbtree.'
\end{itemize}

\subsection{Early Stopping Implementation}
An early stopping mechanism was integrated into the training process to enhance efficiency and avoid overfitting. This mechanism monitored validation performance (10\% of the dataset) and halted training if no improvement in validation loss was observed over five consecutive epochs. This patience threshold balanced adequate time for model convergence with preventing unnecessary computations.

\subsection{Residence 1 Performance Analysis}
\label{sec:res1_analysis}

\begin{figure}[!ht]
    \centering
    \includegraphics[width=0.5\textwidth]{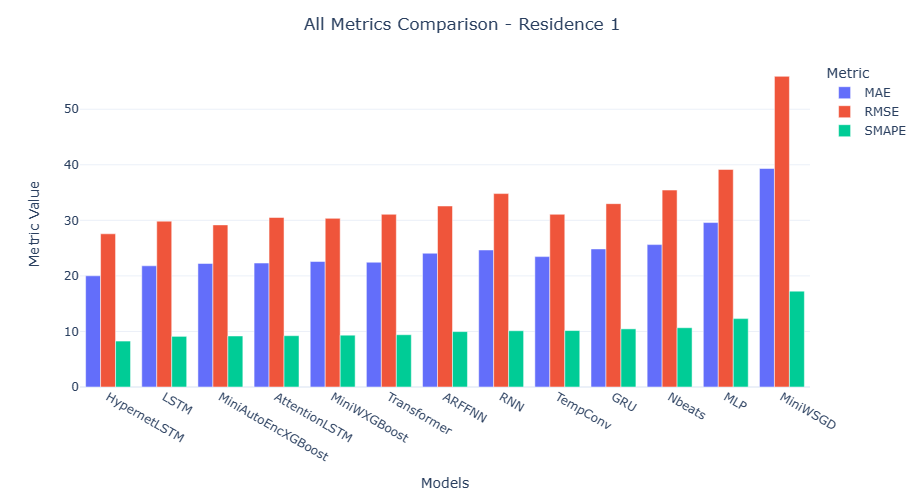}  
    \caption{Comparison of RMSE, SMAPE, and MAE among various models for Residence 1.}
    \label{fig:bar1}
\end{figure}

As shown in Figures~\ref{fig:bar1} and Table~\ref{tab:smape_results}, our proposed HyperNetLSTM model outperforms the competing approaches with the lowest Symmetric Mean Absolute Percentage Error (SMAPE) of 8.87. Notably, it achieves the best Mean Absolute Error (MAE) of 21.03 and a competitive Root Mean Square Error (RMSE) of 29.58. These lower error metrics suggest that the HyperNetLSTM predictions have fewer large deviations, a crucial characteristic in energy forecasting.

\begin{table}[!ht]
    \centering
    \caption{Error metrics for various models, sorted by SMAPE.}
    \begin{tabular}{lrrr}
    \toprule
    \textbf{Model} & \textbf{MAE} & \textbf{RMSE} & \textbf{SMAPE} \\
    \midrule
    HyperNetLSTM       & 21.03 & 29.58 &  8.87 \\
    LSTM               & 21.82 & 29.85 &  9.12 \\
    MiniAutoEncXGBoost & 22.23 & 29.18 &  9.19 \\
    AttentionLSTM      & 22.31 & 30.51 &  9.24 \\
    MiniWXGBoost       & 22.58 & 30.35 &  9.32 \\
    Transformer        & 22.46 & 31.09 &  9.41 \\
    ARFFNN             & 24.08 & 32.58 &  9.97 \\
    RNN                & 24.66 & 34.82 & 10.13 \\
    TempConv           & 23.47 & 31.09 & 10.16 \\
    GRU                & 24.85 & 33.00 & 10.47 \\
    Nbeats             & 25.63 & 35.45 & 10.67 \\
    MLP                & 29.60 & 39.16 & 12.31 \\
    MiniWSGD           & 39.33 & 55.92 & 17.26 \\
    \bottomrule
    \end{tabular}
    \label{tab:smape_results}
\end{table}

\begin{figure}[!ht]
    \centering
    \includegraphics[width=0.5\textwidth]{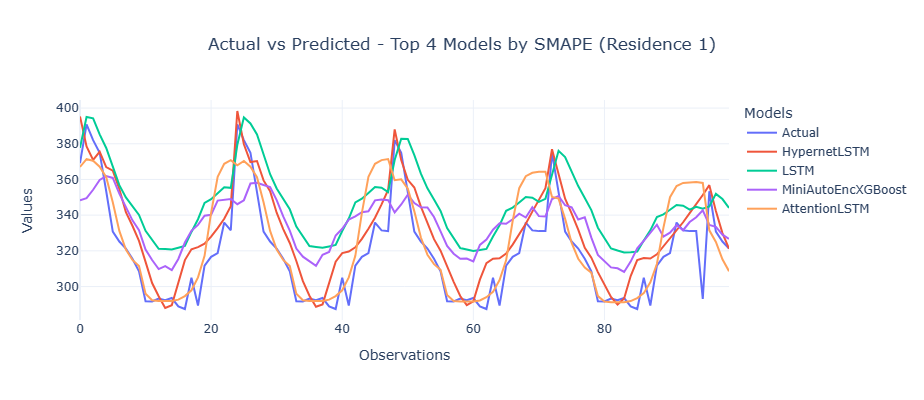}  
    \caption{Comparison of actual energy consumption with predictions by the top four performing models for Residence 1. The line plot illustrates how well each model captures energy usage patterns over several days.}
    \label{fig:line1}
\end{figure}

In contrast, the MiniWSGD model exhibits the highest error metrics, with an MAE of 39.33, RMSE of 55.92, and SMAPE of 17.26, indicating that its predictions deviate significantly from actual values. The strong performance of the HyperNetLSTM is mainly due to LSTM's ability to capture long-term temporal dependencies and the HyperNetwork's dynamic weight adjustment mechanism. Meanwhile, MiniWSGD's poor performance arises from its limited capacity to model the complex temporal patterns inherent in residential energy consumption.

\subsection{Seasonal Performance Analysis}
\label{sec:seasonal_analysis}

We evaluated all models across different seasons for Residence~1 to investigate temporal variations in forecasting accuracy. As shown in Tables~\ref{tab:fig12}, \ref{tab:fig13}, and \ref{tab:fig15}, no single model dominates across all seasons, highlighting the importance of season-specific factors in energy usage.

\begin{table}[!ht]
    \centering
    \caption{Error metrics for various models ordered by SMAPE (Residence 1 -- Fall).}
    \begin{tabular}{lrrr}
    \toprule
    \textbf{Model} & \textbf{MAE} & \textbf{RMSE} & \textbf{SMAPE} \\
    \midrule
    Transformer          & 22.00 & 31.56 &  8.59 \\
    AttentionLSTM        & 24.04 & 33.74 &  9.07 \\
    GRU                  & 25.48 & 33.28 &  9.71 \\
    HyperNetLSTM         & 26.30 & 34.56 &  9.86 \\
    TempConv             & 24.50 & 31.80 &  9.93 \\
    LSTM                 & 28.95 & 38.15 & 11.01 \\
    RNN                  & 32.95 & 44.94 & 11.89 \\
    MiniAutoEncXGBoost   & 32.40 & 38.88 & 12.30 \\
    MiniWXGBoost         & 32.51 & 40.19 & 12.32 \\
    ARFFNN               & 34.49 & 43.91 & 13.05 \\
    Nbeats               & 35.17 & 45.56 & 13.41 \\
    MLP                  & 45.05 & 54.24 & 17.05 \\
    MiniWSGD             & 54.97 & 74.19 & 21.91 \\
    \bottomrule
    \end{tabular}
    \label{tab:fig12}
\end{table}

The Transformer model achieves the best performance in the fall. Its self-attention mechanism helps capture multi-scale temporal features, making it robust to the seasonal fluctuations typical of fall weather.

\begin{table}[!ht]
    \centering
    \caption{Error metrics for various models ordered by SMAPE (Residence 1 -- Winter).}
    \begin{tabular}{lrrr}
    \toprule
    \textbf{Model} & \textbf{MAE} & \textbf{RMSE} & \textbf{SMAPE} \\
    \midrule
    HyperNetLSTM       & 16.44 & 21.08 &  6.11 \\
    MiniAutoEncXGBoost & 16.49 & 20.90 &  6.16 \\
    MiniWXGBoost       & 17.01 & 21.82 &  6.33 \\
    MLP                & 18.68 & 23.35 &  7.00 \\
    LSTM               & 19.27 & 24.81 &  7.25 \\
    ARFFNN             & 19.43 & 24.77 &  7.27 \\
    RNN                & 20.31 & 26.45 &  7.61 \\
    Nbeats             & 23.12 & 30.03 &  8.79 \\
    AttentionLSTM      & 25.52 & 32.02 &  9.65 \\
    TempConv           & 26.53 & 33.42 & 10.09 \\
    Transformer        & 27.73 & 34.40 & 10.40 \\
    GRU                & 29.07 & 36.91 & 11.22 \\
    MiniWSGD           & 29.41 & 38.58 & 11.23 \\
    \bottomrule
    \end{tabular}
    \label{tab:fig13}
\end{table}

During Winter, HyperNetLSTM delivers the best results, underscoring its ability to learn complex consumption patterns exacerbated by heating demands in colder months. MiniAutoEncXGBoost also performs remarkably well, suggesting that autoencoder-based feature extraction combined with gradient boosting is beneficial for winter-specific demand fluctuations.

\begin{table}[!ht]
    \centering
    \caption{Error metrics for various models ordered by SMAPE (Residence 1 -- Summer).}
    \begin{tabular}{lrrr}
    \toprule
    \textbf{Model} & \textbf{MAE} & \textbf{RMSE} & \textbf{SMAPE} \\
    \midrule
    HyperNetLSTM       & 11.99 & 17.73 &  8.13 \\
    Transformer        & 11.89 & 19.20 &  8.42 \\
    AttentionLSTM      & 12.29 & 18.21 &  8.43 \\
    MiniAutoEncXGBoost & 12.85 & 17.46 &  8.56 \\
    GRU                & 12.34 & 18.14 &  8.65 \\
    Nbeats             & 12.61 & 19.51 &  8.74 \\
    MiniWXGBoost       & 13.03 & 18.36 &  8.76 \\
    LSTM               & 13.45 & 19.04 &  9.11 \\
    ARFFNN             & 14.59 & 19.87 &  9.71 \\
    TempConv           & 13.74 & 19.88 &  9.95 \\
    RNN                & 17.96 & 24.98 & 11.98 \\
    MLP                & 21.67 & 26.24 & 13.81 \\
    MiniWSGD           & 26.86 & 37.36 & 18.38 \\
    \bottomrule
    \end{tabular}
    \label{tab:fig15}
\end{table}

In the Summer season, HyperNetLSTM again stands out by accurately modeling abrupt decreases in energy usage during vacation periods. These findings reinforce that specific architectural designs (e.g., HyperNetworks combined with LSTMs) can effectively learn season-specific nuances, while others (such as MiniWSGD) struggle to capture intricate temporal dependencies.

These seasonal insights highlight the need for model selection that aligns with domain-specific characteristics. Depending on the season, distinct models may be more adept at forecasting consumption trends, emphasizing the importance of flexible, context-aware forecasting strategies.

\subsection{Residence 2 Performance Analysis}
\label{sec:res2_analysis}

\begin{figure}[!ht]
    \centering
    \includegraphics[width=0.5\textwidth]{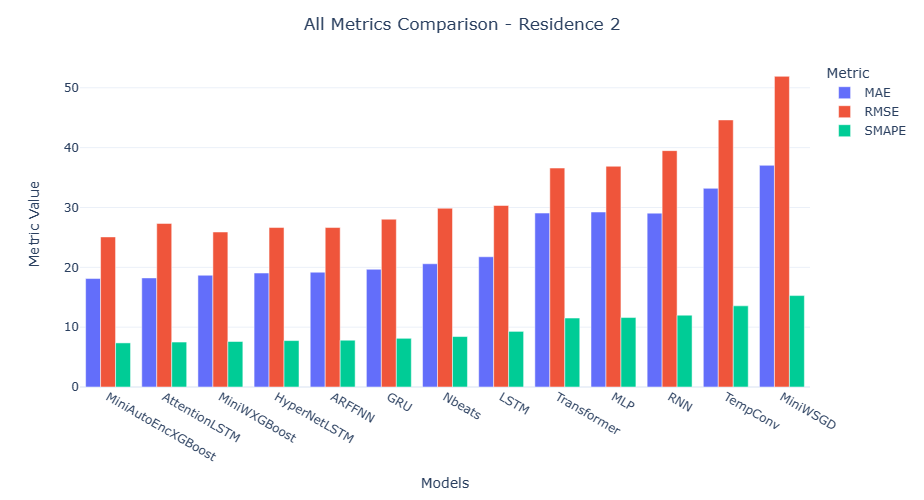}
    \caption{Comparison of RMSE, SMAPE, and MAE among various models for Residence~2.}
    \label{fig:bar2}
\end{figure}

As shown in Table~\ref{tab:fig16} and Figure~\ref{fig:bar2}, the MiniAutoEncXGBoost model achieves the strongest performance for Residence~2, posting the lowest SMAPE of 7.37, MAE of 18.12, and RMSE of 25.08. These metrics underscore its effectiveness in capturing energy usage patterns for this particular household.

\begin{table}[!ht]
    \centering
        \caption{Error metrics for various models ordered by SMAPE for Residence~2.}

    \begin{tabular}{lrrr}
    \toprule
    \textbf{Model} & \textbf{MAE} & \textbf{RMSE} & \textbf{SMAPE} \\
    \midrule
    MiniAutoEncXGBoost & 18.12 & 25.08 &  7.37 \\
    AttentionLSTM      & 18.23 & 27.33 &  7.51 \\
    MiniWXGBoost       & 18.68 & 25.90 &  7.60 \\
    HyperNetLSTM       & 19.04 & 26.65 &  7.76 \\
    ARFFNN             & 19.19 & 26.65 &  7.82 \\
    GRU                & 19.67 & 28.04 &  8.14 \\
    Nbeats             & 20.60 & 29.87 &  8.43 \\
    LSTM               & 21.76 & 30.33 &  9.29 \\
    Transformer        & 29.06 & 36.59 & 11.52 \\
    MLP                & 29.24 & 36.89 & 11.61 \\
    RNN                & 29.05 & 39.50 & 12.00 \\
    TempConv           & 33.20 & 44.63 & 13.56 \\
    MiniWSGD           & 37.04 & 51.91 & 15.28 \\
    \bottomrule
    \end{tabular}
    \label{tab:fig16}
\end{table}

\begin{figure}[!ht]
    \centering
    \includegraphics[width=0.5\textwidth]{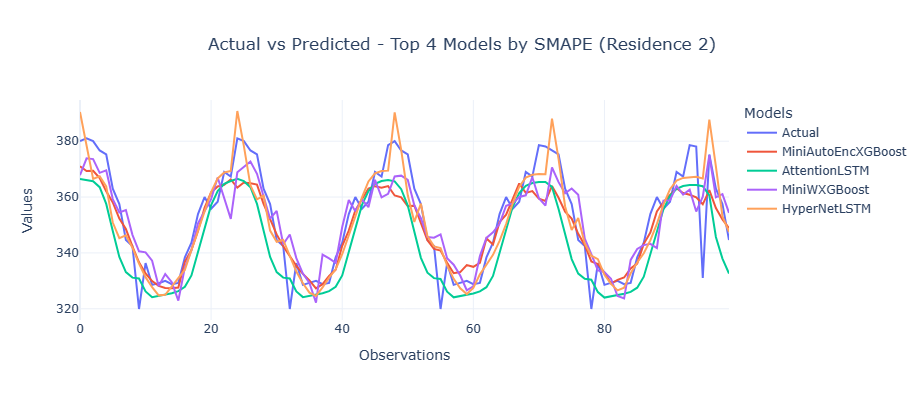}
    \caption{Actual vs.\ predicted energy consumption for Residence~2, highlighting the pattern-learning capabilities of the top four models.}
    \label{fig:line2}
\end{figure}

The AttentionLSTM model also performs competitively, highlighting the utility of attention mechanisms in handling temporal dependencies. By contrast, Transformer-based models show relatively weaker performance for Residence~2, suggesting the need to refine their architectural hyperparameters or training strategies for this dataset. MiniAutoEncXGBoost's superior performance likely stems from the synergy between autoencoder-based feature extraction and gradient boosting, enabling it to uncover intricate consumption patterns.

\subsection{Seasonal Performance Analysis for Residence 2}
\label{sec:res2_seasonal}

We further evaluated all models across different seasons for Residence~2. As shown in Tables~\ref{tab:fig19} through \ref{tab:fig22}, each season presents unique challenges:

- \textbf{Fall (Table~\ref{tab:fig19})}: LSTM achieves the lowest SMAPE of 6.60, indicating its effectiveness at modeling autumn-specific consumption patterns.
- \textbf{Spring (Table~\ref{tab:fig21})}: AttentionLSTM outperforms other methods (SMAPE = 6.80), demonstrating the benefit of integrating attention with LSTM architectures to capture transitions in energy usage.
- \textbf{Summer (Table~\ref{tab:fig22})}: MiniAutoEncXGBoost stands out again with a SMAPE of 12.24, effectively modeling sporadic consumption changes often seen during summer vacations.
- \textbf{Winter (Table~\ref{tab:fig20})}: ARFFNN achieves a SMAPE of 5.60, underscoring its capacity to learn the more pronounced consumption spikes typical in cold weather.

These findings emphasize the importance of selecting models that align well with seasonal demand fluctuations. Adapting to each season's unique consumption characteristics, forecasters can significantly improve prediction accuracy and better manage energy resources.

\begin{table}[!ht]
    \centering
        \caption{Error metrics for various models ordered by SMAPE -- Fall Season for Residence~2}

    \begin{tabular}{lrrr}
    \toprule
    \textbf{Model} & \textbf{MAE} & \textbf{RMSE} & \textbf{SMAPE} \\
    \midrule
    LSTM               & 19.04 & 27.00 &  6.60 \\
    MiniAutoEncXGBoost & 19.61 & 27.06 &  6.78 \\
    GRU                & 19.60 & 26.88 &  6.87 \\
    MiniWXGBoost       & 19.93 & 27.23 &  6.90 \\
    AttentionLSTM      & 20.09 & 28.57 &  6.92 \\
    HyperNetLSTM       & 20.14 & 27.70 &  6.95 \\
    ARFFNN             & 20.79 & 27.38 &  7.23 \\
    Nbeats             & 23.66 & 32.97 &  8.22 \\
    RNN                & 29.19 & 40.23 & 10.03 \\
    TempConv           & 35.42 & 48.81 & 12.09 \\
    MLP                & 35.94 & 43.98 & 12.26 \\
    Transformer        & 35.99 & 43.88 & 12.27 \\
    MiniWSGD           & 42.75 & 58.39 & 15.02 \\
    \bottomrule
    \end{tabular}
    \label{tab:fig19}
\end{table}

\begin{table}[!ht]
    \centering
        \caption{Error metrics for various models ordered by SMAPE -- Winter Season for Residence~2.}

    \begin{tabular}{lrrr}
    \toprule
    \textbf{Model} & \textbf{MAE} & \textbf{RMSE} & \textbf{SMAPE} \\
    \midrule
    ARFFNN             & 12.36 & 15.58 &  5.60 \\
    AttentionLSTM      & 12.10 & 15.47 &  5.62 \\
    MiniAutoEncXGBoost & 12.92 & 16.33 &  5.78 \\
    MiniWXGBoost       & 13.51 & 17.56 &  6.04 \\
    HyperNetLSTM       & 14.00 & 18.26 &  6.29 \\
    GRU                & 15.58 & 21.66 &  7.05 \\
    Nbeats             & 15.60 & 20.50 &  7.05 \\
    LSTM               & 19.09 & 23.68 &  8.93 \\
    Transformer        & 20.67 & 25.85 &  9.17 \\
    MLP                & 20.77 & 26.01 &  9.22 \\
    RNN                & 26.06 & 34.18 & 11.73 \\
    MiniWSGD           & 27.78 & 37.22 & 12.51 \\
    TempConv           & 30.68 & 39.45 & 13.99 \\
    \bottomrule
    \end{tabular}
    \label{tab:fig20}
\end{table}

\begin{table}[!ht]
    \centering
        \caption{Error metrics for various models ordered by SMAPE -- Spring Season for Residence~2.}

    \begin{tabular}{lrrr}
    \toprule
    \textbf{Model} & \textbf{MAE} & \textbf{RMSE} & \textbf{SMAPE} \\
    \midrule
    AttentionLSTM      & 15.48 & 21.73 &  6.80 \\
    MiniAutoEncXGBoost & 16.22 & 21.47 &  7.01 \\
    ARFFNN             & 16.63 & 22.35 &  7.18 \\
    MiniWXGBoost       & 17.09 & 22.96 &  7.37 \\
    Nbeats             & 16.88 & 24.07 &  7.39 \\
    HyperNetLSTM       & 17.48 & 23.79 &  7.50 \\
    GRU                & 17.48 & 23.76 &  7.60 \\
    LSTM               & 21.67 & 27.78 &  9.77 \\
    Transformer        & 23.15 & 29.92 &  9.98 \\
    MLP                & 23.29 & 30.10 & 10.04 \\
    RNN                & 26.20 & 34.83 & 11.70 \\
    TempConv           & 32.75 & 42.19 & 14.27 \\
    MiniWSGD           & 34.13 & 47.84 & 15.22 \\
    \bottomrule
    \end{tabular}
    \label{tab:fig21}
\end{table}

\begin{table}[!ht]
    \centering
        \caption{Error metrics for various models ordered by SMAPE -- Summer Season for Residence~2.}
    \begin{tabular}{lrrr}
    \toprule
    \textbf{Model} & \textbf{MAE} & \textbf{RMSE} & \textbf{SMAPE} \\
    \midrule
    MiniAutoEncXGBoost & 27.81 & 36.64 & 12.24 \\
    MiniWXGBoost       & 28.45 & 37.76 & 12.53 \\
    HyperNetLSTM       & 28.85 & 39.03 & 12.74 \\
    Nbeats             & 28.55 & 42.40 & 12.83 \\
    AttentionLSTM      & 29.95 & 44.10 & 13.42 \\
    GRU                & 30.58 & 42.53 & 13.58 \\
    ARFFNN             & 32.68 & 42.63 & 14.29 \\
    TempConv           & 34.26 & 48.06 & 14.98 \\
    LSTM               & 33.00 & 47.03 & 15.10 \\
    Transformer        & 38.40 & 44.53 & 16.47 \\
    MLP                & 39.23 & 45.54 & 16.85 \\
    RNN                & 37.99 & 51.17 & 16.96 \\
    MiniWSGD           & 47.48 & 65.99 & 21.58 \\
    \bottomrule
    \end{tabular}
    \label{tab:fig22}
\end{table}

\section{Conclusion}
This study's detailed analysis across various seasons and techniques confirms that no single model performs best in all scenarios. Seasonal variations influence energy forecasting accuracy and require seasonally contextualized model selection to handle the unique characteristics of energy consumption patterns effectively. Additionally, the variability introduced by human behavior poses significant challenges for forecasting models. Current standalone AI models may not fully capture these complexities. We propose developing innovative human-AI hybrid models integrating human-centric considerations alongside seasonal factors. Such models would account for the unpredictable nature of human activity while improving forecasts' responsiveness to fluctuating energy demand patterns. This approach represents a promising direction for future research, offering the potential to significantly enhance the precision and reliability of energy forecasting models in residential settings.

\bibliographystyle{IEEEtran}  
\bibliography{ref}
\end{document}